\documentclass[conference]{IEEEtran}
\usepackage{cite}
\usepackage{amsmath,amssymb,amsfonts}
\usepackage[linesnumbered, ruled, vlined]{algorithm2e}
\usepackage{graphicx}
\usepackage{textcomp}
\usepackage{xcolor}
\usepackage[font=small,labelfont=bf]{caption}
\usepackage{hyphenat}
\usepackage{tikz}
\usepackage{pgfplots}
\usepackage{amssymb}
\usepackage{mathtools}
\usepackage{array}
\usepackage{hyperref} 
\usepackage{subcaption}
\usepackage[linesnumbered,ruled,vlined]{algorithm2e}
\usepackage{balance}
\usepackage{xcolor}
\usepackage{tikz}
\usepackage[utf8]{inputenc}
\usepackage[noend]{algpseudocode}
\usepackage{float} % For generating filler text
\usepackage{colortbl}
\usepackage{algorithmicx}
\usepackage{lipsum}

\hypersetup{
    colorlinks=true,
    linkcolor=blue,
    citecolor=blue,
    urlcolor=blue
}
\usepackage{graphics}
\pgfplotsset{compat=1.18}
\begin{document}
% \title{FLamma: Adaptive Gamma-Based Game Theoretic Framework for Fair Federated Learning }

% \title{FLAMEGAME: Federated Learning with Adaptive Mechanisms and Equilibrium in Game Theory}
\title{Incentive-Compatible Federated Learning with Stackelberg Game Modeling }

% \author
% {Anonymous Authors}
\author{
 Simin Javaherian~~~~~~Bryce Turney~~~~~~Li Chen~~~~~~ Nian-Feng Tzeng\\
% \IEEEauthorblockA{
 School of Computing and Informatics,
University of Louisiana at Lafayette \\
Email: \{simin.javaherian1, bryce.turney1, li.chen, nianfeng.tzeng\}@louisiana.edu }
% }
\maketitle
\begin {abstract}
Federated Learning (FL) has gained prominence as a decentralized machine learning paradigm, allowing clients to collaboratively train a global model while preserving data privacy. Despite its potential, FL faces significant challenges in heterogeneous environments, where varying client resources and capabilities can undermine overall system performance. Existing approaches primarily focus on maximizing global model accuracy, often at the expense of unfairness among clients and suboptimal system efficiency, particularly in non-IID (non-Independent and Identically Distributed) settings. In this paper, we introduce {\em FLamma}, a novel \underline{F}ederated \underline{L}earning framework based on adaptive g\underline{amma}-based Stackelberg game, designed to address the aforementioned limitations and promote fairness. Our approach allows the server to act as the leader, dynamically adjusting a decay factor while clients, acting as followers, optimally select their number of local epochs to maximize their utility. Over time, the server incrementally balances client influence, initially rewarding higher-contributing clients and gradually leveling their impact, driving the system toward a Stackelberg Equilibrium. Extensive simulations on both IID and non-IID datasets show that our method significantly improves fairness in accuracy distribution without compromising overall model performance or convergence speed, outperforming traditional FL baselines.
\end{abstract}
\section{Introduction}\label{Introduction}

Federated learning (FL)\cite{mcmahan2017communication} has emerged as a promising approach to address the challenges of centralized machine learning by enabling collaborative model training across distributed devices while preserving data privacy. This decentralized paradigm allows for the efficient utilization of distributed data resources while mitigating privacy concerns associated with centralized data storage and processing. The impact of FL is evident in numerous collaborative studies, such as clustering \cite{islam2024fedclust, ghosh2020efficient}, client selection \cite{cho2020client, javaherian2024fedfair, lai2021oort, ezzeldin2023fairfed}, and personalization \cite{mansour2020three, li2021ditto}. However, the assumption that clients are inherently motivated to participate actively is often made. To establish an effective FL system, it is crucial to attract and retain a large number of participating clients \cite{blum2021one}. Consequently, clients need to be incentivized individually to participate in training. Therefore, designing effective incentive mechanisms is essential to ensure the active and beneficial participation of all clients.

This is where game theory \cite{maschler2020game} and incentive mechanisms become crucial. By leveraging game theory, we can tailor incentives that align individual client objectives with the collective goals of the learning network. This involves formulating a payoff structure that rewards clients based on multiple factors, including their accuracy and consistency in contributing to the global model. Such mechanisms help mitigate the risks of free-riding and data hoarding, encouraging clients to contribute high-quality data and computational resources. Furthermore, these game-theoretic strategies can dynamically adjust to varying network conditions and client capabilities, ensuring a robust and adaptive FL system that optimizes both individual and collective outcomes.

However, previous works in game theory and incentive mechanism designs for FL have several limitations. Many existing studies primarily focus on optimizing immediate monetary rewards or simplistic participation incentives, offering fixed payments for data points without considering the nuanced contributions of each client \cite{khan2020federated, sarikaya2019motivating}. These models often overlook the complex dynamics of client contributions. Furthermore, traditional models generally do not consider long-term fairness, and they fail to account for the temporal aspects of client engagement, which can vary dramatically across different phases of the learning process.

Our approach, titled \textit{FLamma}, addresses the shortcomings of existing methods by incorporating a more nuanced understanding of client behavior and resource allocation. FLamma leverages a Stackelberg game-theoretic framework \cite{amir1999stackelberg}, where clients choose their number of local epochs based on their utility maximization, while the server dynamically adjusts a decay factor to regulate client contributions over time. We introduce a novel utility function with dynamic allocation coefficients that adjust the influence of individual contributions over time. This adjustment promotes a more equitable distribution of training time and aligns with the long-term objectives of sustaining high-quality model performance and ensuring system robustness. By incorporating a decay factor into the assignment strategy, our model accounts for temporal variations in client engagement, promoting fairness while maintaining convergence and performance over time. This approach addresses both the immediate and evolving nature of client contributions, ensuring a more balanced and efficient resource allocation. Through experimental validation, we demonstrate the effectiveness of our approach in addressing these challenges and enhancing the performance of FL systems. 
 FLamma achieves an impressive improvement in accuracy by 120.93\% over q-FFL and reduces variance by 99.09\% compared to FedAvg, demonstrating its effectiveness in both accuracy and fairness. As a result, our method enhances the robustness and flexibility of FL systems, effectively handling the diverse capabilities and resource availability of clients.

\noindent We summarize our \underline{contributions} as follows: \begin{itemize}

\item We model the interaction between the server and clients using a Stackelberg game-theoretic framework, where the server dynamically adjusts a decay factor, and clients optimize their local epochs to maximize utility. This approach achieves Stackelberg Equilibrium, promoting fairness, encouraging active participation, and enhancing system stability and efficiency.

\item The decay factor adapts to temporal variations in client engagement, preventing over-contribution by certain clients and ensuring balanced participation throughout the training process, which fosters a fair and collaborative learning environment.

\item We provide experimental validation results showing that our method achieves a fairer accuracy distribution across clients while maintaining strong global model performance. Our approach outperforms traditional FL methods in both fairness and overall system performance. 

\item We provide a convergence analysis of our proposed method, demonstrating that it maintains competitive convergence rates, ensuring timely model updates without compromising accuracy or fairness.

\end{itemize}

\section{Background and Related works}\label{relatedworks}

In the realm of federated learning, numerous studies have explored various strategies to enhance collaborative model training under different settings. %Additionally, research has focused on designing incentive mechanisms to improve client participation and overall system efficiency. 
In this section, we review related work from the following key aspects: incentive mechanisms, client selection and fairness.

\vspace{1em}
\noindent \textit{\textbf{Incentive Mechanisms.}} A substantial body of research has focused on designing incentive mechanisms to promote client participation in FL, using approaches such as game theory, contract theory, and reputation-based systems.

\begin{itemize}
    \item {\em Game theory} is commonly applied to optimize contributions in both cooperative and competitive settings. For example, Zhang et al. \cite{zhang2024incentivizing} proposed a Stackelberg model where rewards are allocated based on model quality, addressing data quality variance in FL. Wu et al. \cite{wu2023incentive} introduced a Stackelberg game using importance sampling to incentivize clients based on their contributions. Dorner et al. \cite{dorner2024incentivizing} focused on incentivizing honest client updates in collaborative learning by discouraging dishonest behavior. Additionally, non-cooperative games have been explored, such as Khawam et al. \cite{khawam2024non}, who modeled edge selection as a non-cooperative game where IoT devices autonomously minimize learning errors and communication costs. 
Furthermore, coalition games have been used to encourage clients to form coalitions that minimize costs and maximize collective benefits, addressing data heterogeneity and variance-bias trade-offs \cite{donahue2021model, zhang2024coalitional}. Yu et al. \cite{yu2020fairness} proposed the Federated Learning Incentivizer (FLI) scheme, which dynamically allocates budgets among data owners to optimize their utility. Hu et al. \cite{hu2024game} introduced a quality-aware incentive mechanism for hierarchical FL using a three-layer Stackelberg game (cloud-edge-en) to optimize cooperation. Training-time incentives were explored by Wu et al. \cite{wu2024incentive}, who proposed an algorithm offering real-time model rewards based on client contributions. Durand et al. \cite{durand2024federated} and Arisdakessian et al. \cite{arisdakessian2023coalitional} applied hedonic games to minimize learning errors and communication costs, while Khawam et al. \cite{khawam2023edge} examined coalition formation among LoRaWAN devices. Hasan et al. \cite{hasan2021incentive} explored stable coalitions in FL using hedonic games and Nash-stable partitions. Murhekar et al. \cite{murhekar2024incentives} introduced a budget-balanced mechanism to maximize collective welfare at Nash equilibrium, and IncFL \cite{cho2022federate} dynamically adjusts aggregation weights to incentivize more clients to adopt the global model.

\item {\em Contract theory} has been used in some works for FL environments where the server lacks full visibility into client resources or data quality \cite{liu2022contract}. Ding et al. \cite{ding2020optimal} introduced a contract-theoretic mechanism optimizing user participation based on private information like training costs and communication delays. Zhan et al. \cite{zhan2020incentive} developed a multidimensional contract model balancing computation latency, communication time, and payment. Yu et al. \cite{yu2024contract} proposed a contract-based incentive mechanism for wireless networks, maximizing server utility while ensuring individual rationality and incentive compatibility. These methods are particularly effective in cross-silo FL, where organizations cooperate to maximize social welfare \cite{tang2021incentive}.

\item {\em Reputation-based} systems reward clients based on historical performance, promoting trust and participation. Kang et al. \cite{kang2019incentive} combined reputation systems with contract theory to ensure reliable mobile device participation.

\item {\em Reinforcement learning (RL)} has also been used in FL incentive mechanisms. Zhan et al. \cite{zhan2020learning} designed a deep RL-based mechanism for optimizing pricing and training strategies, while Shen et al. \cite{shen2024federated} introduced a DRL-based approach optimizing client satisfaction, resource allocation, and faster convergence.
\end{itemize}
%Comprehensive reviews by Zeng et al. \cite{zeng2021comprehensive, zeng2022incentive} and Ali et al. \cite{ali2023systematic} provide in-depth analyses of the various incentive mechanisms used in FL, covering approaches like Stackelberg games, auctions and contract theory. 

\noindent \textit{\textbf{Client selection and fairness.}} Client selection in FL has been one of the major focuses of existing efforts, aiming to balance performance, efficiency, and fairness in the system. Various approaches have been introduced to ensure that clients with diverse data and resources are fairly represented in the training process. For instance, Donahue et al. \cite{donahue2023fairness} examined two types of fairness in federated learning: egalitarian fairness, which aims to limit the disparity in error rates across clients, and proportional fairness, which seeks to reward clients based on their data contributions, ensuring that those who contribute more data benefit proportionally. There are also some fairness-driven algorithms, such as FairFedCS \cite{shi2023fairness}, which dynamically adjust client selection probabilities by considering factors like reputation, participation, and contributions, ensuring fair treatment while maintaining model performance. AdaFL \cite{li2024adafl} introduces a dynamic client selection strategy that adjusts the number of participating clients over time, starting with a smaller selection to reduce communication overhead and progressively increasing it to improve model generalization. Some fair client selection methods, such as FedFair$^3$ \cite{10622273}, prioritize clients and adjust their participation to reduce accuracy variance, while Eiffel \cite{sultana2022eiffel} focuses on resource efficiency by adaptively selecting mobile devices and adjusting update frequencies to ensure fairness in federated learning. 
q-FFL \cite{li2019fair} and AdaFed \cite{hamidi2024adafed} promote fairness by prioritizing clients with higher loss values. q-FFL introduces an optimization objective that encourages a more uniform accuracy distribution across devices, while AdaFed dynamically adjusts the shared update direction by considering both local gradients and loss functions, ensuring clients with larger loss values see greater improvements, thereby promoting a fairer learning process.\\

\noindent{\bf \textit{Motivation}}: Previous works typically focus on either allocating resources based on client contributions or distributing resources equally among clients. However, a significant gap remains in addressing both aspects simultaneously—balancing fairness with efficiency. Our model fills this gap by integrating these two approaches through the use of a \textit{decay factor}. Initially, our approach gives greater weight to clients who achieve high contributions, allowing higher-contributing clients to have a larger influence on the global model. Over time, this influence is gradually reduced, ensuring that all clients have equal training opportunities. Clients independently choose their number of local epochs to maximize their utility, while the server adjusts the decay factor dynamically to maintain fairness across participants. This dynamic adjustment encourages a more balanced and fair resource allocation, which is essential for both system fairness and efficiency. By incorporating a decay factor into the assignment strategy, our model adapts to temporal variations in client engagement and their epoch choices, promoting sustained fairness and convergence as training progresses. This results in a more balanced and efficient allocation of resources throughout the learning process, ensuring robust system performance and improved outcomes for all participants.

 \section{Preliminaries}

\subsection{Federated Learning}

The primary objective in FL is to minimize the global loss across all participating clients. This global loss function, denoted as \( F(w) \), aggregates the local loss functions \( F_i(w) \) computed on the individual datasets of each client. The objective function is expressed as:

\[
\arg\min_{w} F(w) = \sum_{i=1}^N p_i F_i(w),
\]
where \( F_i(w) \) represents the local loss function for client \( i \), and \( p_i \) is the weight associated with each client, calculated as \( p_i = \frac{|D_i|}{|D|} \), where \( |D_i| \) is the number of data points owned by client \( i \), and \( |D| \) is the total number of data points across all clients. 

\subsection{Game Theory and Stackelberg Games}
Game theory\cite{maschler2020game} provides a formal framework for modeling interactions between rational agents (players) whose decisions influence one another's outcomes. Each player selects from a set of strategies, and the resulting payoff depends on the combination of strategies chosen by all participants. The utility of any player \(i\) is generally expressed as:
\[
U_i = R_i - C_i,
\]
where \(R_i\) represents the reward or benefit derived from a chosen strategy, and \(C_i\) denotes the associated cost. 

A Stackelberg game is a hierarchical type of leader-follower game, involving a \textit{leader} and one or more \textit{followers}. 
% The leader moves first, selecting a strategy, and the followers react by optimizing their strategies in response. 
\begin{itemize}
    \item \textbf{Leader:} The player who moves first, following the strategy denoted by \(\gamma\).
    \item \textbf{Followers:} A set of players who observe the leader’s action and respond with strategies \(\tau_i = \{\tau_1, \tau_2, \dots, \tau_N\}\).
\end{itemize}

\textit{Leader's Utility:}
The leader's utility function \( U_{\text{server}}(\gamma, \tau_i) \) depends on the leader’s strategy \(\gamma\) and the followers’ strategies \(\tau_i\). The leader aims to maximize its utility:
\[
\max_{\gamma} U_{\text{server}}(\gamma, \tau_i^*),
\]
where \( \tau_i^* \) represents the optimal responses of the followers, given the leader’s strategy \(\gamma\).

 \textit{Follower's Utility:} Given the leader's strategy $\gamma$,
each follower \(i\) has its own utility function \( U_i( \gamma, \tau_i, \tau_{-i}) \) with their own strategy \(\tau_i\) and other clients strategy $\tau_{-i}$. Each follower \(i\) solves:
\[
\max_{\tau_i} U_i(\gamma, \tau_i, \tau_{-i}).
\]
\textit{Stackelberg Equilibrium} is achieved when the leader's strategy maximizes its utility while accounting for the optimal responses of the followers.
\begin{figure}[h]
\centering
{\includegraphics[width=0.5\textwidth]{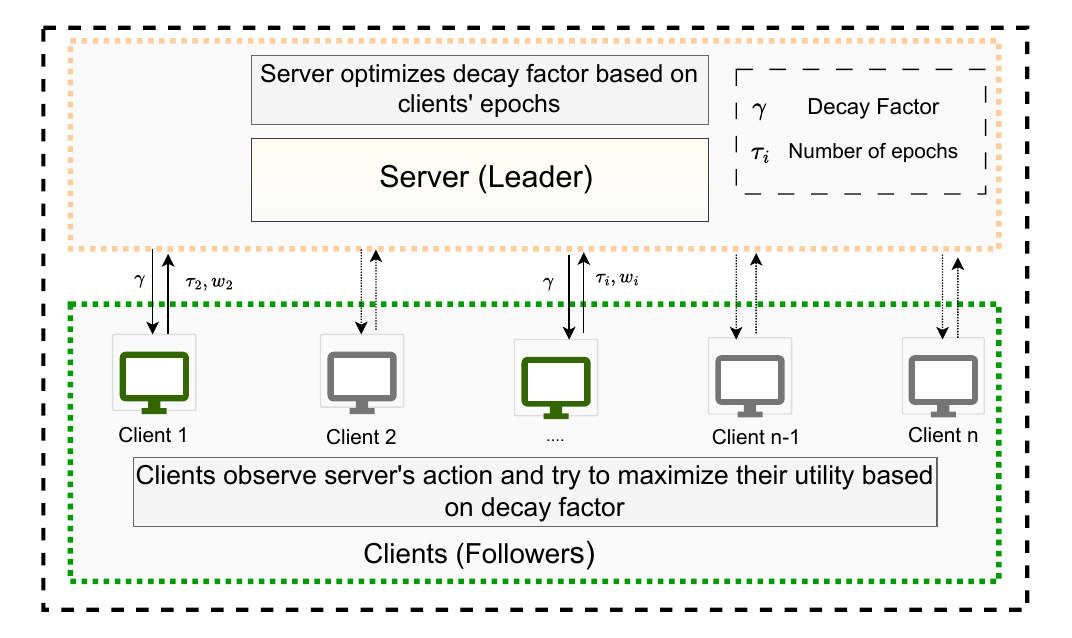}}
\caption{Overview of {\em FLamma}, an adaptive gamma-based game theoretic
framework for fair federated learning.}
\label{figure1}
\end{figure}
\section{Design of {\em FLamma}}\label{Methodology}

This section introduces the design of our game-theoretic framework for fair federated learning. 

\subsection{Overview}
{\em FLamma} leverages a Stackelberg game-theoretic framework to dynamically adjust client contributions in an FL environment. The game is designed to achieve equilibrium by leveraging a decay factor that the server adjusts to modulate the effective contributions of each client over time. The server’s role is to maximize efficiency and fairness by dynamically adjusting the influence of each client over time through the decay factor \(\gamma\), which ensures balanced updates across clients.
Initially, all clients contribute based on their local resources and data quality, but as the process evolves, the decay factor reduces the influence of individual contributions. This adjustment helps guide the system toward the Stackelberg Equilibrium, where clients and server both optimize their strategies. Fig. \ref{figure1} shows a general overview of our setting, where clients are selected based on their contribution and their contribution influence gets adjusted over time. %The green clients show selected clients for a specific round. 
As shown in the figure, the server sends the decay factor $\gamma$ to the clients and selected clients (in green) send their chose number of training epochs, $\tau_i$, as well as their local updates, in response to the server.

% The server’s role is to minimize this global loss \(F(w)\) by dynamically adjusting the influence of each client over time through the decay factor \(\gamma\), which ensures balanced updates across clients.

\subsection{Game Strategy}

In our FL framework, the interaction between the server and clients is framed as a Stackelberg game, where the server acts as the leader and the clients are the followers. The server commits to a strategy by adjusting the decay factor \( \gamma \), while the clients respond optimally by selecting the number of local epochs \( \tau_i \) that maximizes their utility. Both the server's and clients' utility functions are defined in a general form as the difference between revenue and cost.
% \begin{equation}
%  \nonumber  U_i=R_i-C_i. 
% \end{equation}

\vspace{1em}

\noindent\textit{Server Utility.} The server’s utility is defined as:
\[
\begin{aligned}
U_{\text{server}}(\gamma, \tau_i) &= \sum_{i=1}^N \left[ \gamma \cdot \left((1 - \frac{\| w_i^t - w^t \|}{\| w^t \|} ) + \tau_i \right) \right] - t \cdot \gamma^2, \\
&\text{s.t: } 0 \leq \gamma \leq 1
\end{aligned}
\]
where \( \gamma \) is the decay factor, \( w_i^t \) is the local model of client \( i \), \( w^t \) is the global model at round \( t \), \( \tau_i \) is the number of local epochs for client \( i \), \( t \) is the number of global rounds.  
\( \omega_i = 1 - \frac{\| w_i^t - w^t \|}{\| w^t \|} \) is the clients' contribution which captures the discrepancy between the client’s local model and the global model.

\vspace{1em}

\noindent\textit{Client Utility.} The utility function for each client \( i \) is defined as:
\begin{equation}
  U_i(\gamma, \tau_i, \tau_{-i}) = \gamma \cdot \left( 1 - \frac{\| w_i^t - w^t \|}{\| w^t \|} \right) \cdot \tau_i - c_i \tau_i^2  
  \label{eq1}
\end{equation}
where \( U_i(\gamma, \tau_i, \tau_{-i}) \) is the utility of client \( i \) when it uses strategy \( \tau_i \) and the other clients use strategies \( \tau_{-i} \),
 \( \tau_i \) represents the number of local epochs chosen by client \( i \).

 \vspace{0.8em}

\noindent\textbf{\textit{Definition 1.}} \textit{Individual Rationality (IR)}. In our FL game, the IR constraint ensures that each client’s utility is non-negative, meaning that participating in the learning process should provide a utility greater than or equal to 0. Mathematically, the IR constraint is defined as:
\[
U_i(\gamma, \tau_i, \tau_{-i}) \geq 0 \quad \forall i.
\]
For client \( i \) to participate in the federated learning process, its utility must be non-negative, ensuring that it has an incentive to follow the strategy that maximizes its utility. This constraint guarantees that all clients benefit from participating in the collaborative learning process.

\vspace{0.6em}
\subsection{Stackelberg Equilibrium}

In the Stackelberg game, the server commits to the decay factor \( \gamma \), and the clients respond optimally by selecting the number of local iterations \( \tau_i^* \). We first compute the clients' optimal \( \tau_i^* \), then use it in the server utility function to calculate the optimal \( \gamma^* \).

To find the optimal number of local iterations \(\tau_i^*\), we first compute the derivative of the utility function with respect to \(\tau_i\):
\[
\frac{d}{d\tau_i} U_i(\gamma, \tau_i, \tau_{-i}) = \gamma \cdot \left( 1 - \frac{\| w_i^t - w^t \|}{\| w^t \|} \right) - 2 c_i \tau_i.
\]
Setting the derivative equal to zero and solving it for \( \tau_i \), we find the optimal local iterations for client \( i \) as:
\[
\tau_i^* = \frac{\gamma \cdot \left( 1 - \frac{\| w_i^t - w^t \|}{\| w^t \|} \right)}{2 c_i}.
\]

\noindent\textit{\textbf{Lemma 1.}} The client utility function \( U_i(\gamma, \tau_i, \tau_{-i}) \) is concave with respect to the number of local epochs \( \tau_i \).

\noindent \textit{Proof.} Given the client utility by Eq.~(\ref{eq1}), to check for concavity, we compute the second derivative of \( U_i \) with respect to \( \tau_i \).
Taking the second derivative, we have:
\[
\frac{d^2}{d\tau_i^2} U_i(\gamma, \tau_i, \tau_{-i}) = -2 c_i.
\]
Since \( c_i > 0 \), the second derivative is negative, confirming that the utility function \( U_i \) is concave with respect to \( \tau_i \).

% Taking the second derivative confirms concavity:
% \[
% \frac{d^2}{d\tau_i^2} U_i(\tau_i, \tau_{-i}) = -2 c_i.
% \]
 % \vspace{1em}
\vspace{0.8em}
\noindent\textbf{\textit{Lemma 2.}} Client Utility is maximized at the optimal number of local epochs \( \tau_i^* \).

\noindent\textit{Proof.} Since \( -2 c_i \) is negative, the utility function is concave, meaning that \( \tau_i^* \) is a maximum. 
 % \vspace{1em}
\vspace{0.8em}

\noindent\textbf{\textit{Definition 2.}} \textit{Best Response.} The best response for a client \( i \) is the strategy \( \tau_i \) that maximizes their utility given the strategies of other clients \( \tau_{-i} \). Mathematically, the best response for client \( i \) is:
\[
B_i(\tau_{-i}) = \arg\max_{\tau_i} U_i(\gamma, \tau_i, \tau_{-i}).
\]
In this game, the optimal strategy \( \tau_i^* \) for each client is the best response to the server’s decay factor \( \gamma \) and the other clients’ strategies.

Now, we substitute \( \tau_i^* \) into the server's utility function:
\[
U_{\text{server}}(\gamma, \tau_i) = \gamma \cdot (1 - \frac{\| w_i^t - w^t \|}{\| w^t \|} ) + \frac{\gamma^2 \cdot (1 - \frac{\| w_i^t - w^t \|}{\| w^t \|} )}{2 c_i} - t \cdot \gamma^2.
\]
% Where \( C_i = 1 - \frac{\| w_i^t - w^t \|}{\| w^t \|} \).
To find \( \gamma^* \), we take the derivative of the server utility function with respect to \( \gamma \):
\[
\frac{d}{d\gamma} U_{\text{server}}(\gamma, \tau_i) = (1 - \frac{\| w_i^t - w^t \|}{\| w^t \|} ) + \frac{\gamma \cdot (1 - \frac{\| w_i^t - w^t \|}{\| w^t \|} )}{c_i} - 2 t \cdot \gamma.
\]
Setting the derivative equal to zero and solving it for gamma, we find the \( \gamma^* \) as:
\[
\gamma^* = \frac{(1 - \frac{\| w_i^t - w^t \|}{\| w^t \|} ) \cdot c_i}{2 t \cdot c_i - (1 - \frac{\| w_i^t - w^t \|}{\| w^t \|} )}.
\]
% \[
% C_i + \frac{\gamma \cdot C_i}{c_i} = 2 t \cdot \gamma
% \]

% Solve for \( \gamma^* \):

% \vspace{1em}

\noindent\textbf{\textit{Lemma 3.}} Server Utility is maximized at $\gamma^*$.

\noindent\textit{Proof.} Since the second derivative is negative for all $\gamma$, $U_{\text{server}}(\gamma, \tau_i)$ is a concave function. Therefore, the optimal $\gamma^*$ maximizes the utility function.

% \vspace{1em}
\vspace{0.8em}
\noindent\textbf{\textit{Definition 3.}} \textit{Nash Equilibrium}. A Nash equilibrium is a strategy profile \( \tau^* = (\tau_1^*, \dots, \tau_N^*) \) where no client \( i \) can increase their utility by unilaterally deviating from their strategy. This implies that:
   \[
   U_i(\gamma, \tau_i^*, \tau_{-i}^*) \geq U_i(\gamma, \tau_i, \tau_{-i}^*) \quad \forall \tau_i, i.
   \]

% \vspace{1em}

\begin{algorithm}[h]
\caption{FLamma Algorithm}\label{FLaim}
\SetKwInOut{Input}{Input}
\SetKwInOut{Output}{Output}

\Input{Initial global model parameter $w_0$, total rounds $T$, local epochs $\tau_i$, learning rate $\eta$}
\Output{Final global model parameter $w_T$}
\BlankLine

\For{each round $t = 1, 2, \ldots, T$}{
    \textbf{Server-side:}\\ 
    \Indp Calculate contribution $\omega_i$ for each client $i$\;
    Select a subset of clients $S_t$ based on contributions $\omega_i$\;
    Broadcast global model parameter $w_t$ and decay factor $\gamma$ to the selected clients\;
    
    \Indm \For{each client $k \in S_t$ \textbf{in parallel}}{
        \textbf{Client-side (Local Training):}\\
        \Indp Initialize local model $w_t^k \gets w_t$\;
        Client optimally selects the number of local epochs $\tau_k$ to maximize utility\;
        
        \For{each local epoch $e = 1, 2, \ldots, \tau_k$}{
            Perform mini-batch gradient descent on local data\;
            Update local model: $w_t^k \gets w_t^k - \eta \gamma \nabla F_k(w_t^k)$\;
        }
        Send updated local model $w_t^k$ back to the server\;
        \Indm
    }
    
    \textbf{Server Aggregation:}\\
    \Indp Aggregate the received local models: $w_{t+1} \gets \sum_{k \in S_t} p_k w_t^k$\;
    Calculate utility $U_i$ for each client $i$\;
    Update the decay factor $\gamma$\;
    %$\gamma_i(t) = \frac{1}{1 + \lambda (t+\tau_i)}$\;
    \Indm
}
\textbf{Output:} Final global model parameter $w_T$\;
\end{algorithm}

Given this context, we summarize the steps of Algorithm \ref{FLaim} as follows:

\begin{itemize}
    \item  \textbf{Step 1.} Server calculates the client contributions and selects clients based on their contributions. It then sends the model to the selected clients (Lines 1-5).
\item \textbf{Step 2.} In parallel, each selected client performs local training for the determined number of epochs. Updated local model is then sent to the server (Lines 6-13).

\item \textbf{Step 3.} Server aggregates the local model updates received from the clients (Lines 14-15).
\item \textbf{Step 4.} Server calculates the utility for each client based on their contributions and local epochs and then updates the decay factor (Lines 16-17).

\item \textbf{Step 5.} The algorithm outputs the final global model at the end of the rounds (Lines 18).

\end{itemize}
\vspace{0.6em}
\noindent\textbf{\textit{Lemma 4.}} \textit{The client sub-game has at least one Nash equilibrium.}

\noindent\textit{Proof.} Each client's utility function is given by Eq.~(\ref{eq1})
where \( \tau_i \) represents client \( i \)'s number of local epochs, and \( \tau_{-i} \) denotes the strategies of the other clients.
Given that $\tau_i^*$ is client \( i \)'s best response \( B_i(\tau_{-i}) \) and the strategy space \( \tau_i \in [\tau_{\min}, \tau_{\max}] \) is compact and convex, and using lemma 1, \( U_i(\gamma, \tau_i, \tau_{-i}) \) is concave in \( \tau_i \), by Fixed Point Theorem\cite{rosen1965existence}, a continuous function from a compact, convex set to itself has a fixed point. The best response mappings \( B_i(\tau_{-i}) \) define such a function, ensuring the existence of a fixed point \( \tau^* = (\tau_1^*, \dots, \tau_N^*) \). This fixed point represents a Nash equilibrium, where each client’s strategy is a best response to the others, meaning no client can increase their utility by unilaterally deviating from their current strategy.

To demonstrate the existence of a Nash equilibrium, consider that each client \( i \) has a unique best response \( \tau_i^* \), while all other clients \( j \neq i \) follow their best responses \( \tau_j^* \). Since no client can improve their utility by deviating from their best response \( \tau_i^* \), the strategy profile \( \tau^* = (\tau_1^*, \dots, \tau_N^*) \) constitutes a Nash equilibrium for the client sub-game. Therefore, the client sub-game admits at least one Nash equilibrium.

\vspace{0.6em}

\noindent\textbf{\textit{Lemma 5.}} The equilibrium of the clients’ level sub-game always satisfies the IR constraint.

\noindent\textit{Proof.} Consider an equilibrium \( \tau^* \), and let \( i \) be a client such that: $U_i(\gamma,\tau_i^*, \tau_{-i}^*) < 0.$ Now, substitute \( \tau_i^* \) with \( 0 \):
\[
U_i(\gamma, 0, \tau_{-i}^*) = 0 > U_i(\gamma,\tau_i^*, \tau_{-i}^*).
\]
This contradicts the definition of Nash equilibrium. Therefore, we have: $U_i(\gamma,\tau_i^*, \tau_{-i}^*) \geq 0 \quad \text{for all } i.$

% \vspace{1em}

\vspace{0.6em}

\noindent\textbf{\textit{How does FLamma respect fairness?}} As training progresses, \( \gamma \) decays, gradually diminishing the influence of high-contributing clients and giving more weight to clients with lower contributions. By reducing the disparity in client influence, \( \gamma \) helps to balance out the updates from different clients, leading to a more uniform distribution of accuracy. This dynamic adjustment helps mitigate the risk of overfitting to data from dominant clients, ensuring that all clients, regardless of their initial contribution levels, have a fair chance to contribute to the global model. As a result, the variance in accuracy across clients is reduced, leading to a more equitable model that performs consistently well for all participants. By fine-tuning client contributions over time, \( \gamma \) helps ensure that high-performing clients do not dominate the global model, thus maintaining fairness in accuracy distribution. This not only promotes long-term equity but also stabilizes the system, making the overall learning process more robust and improving the quality of the final global model across all clients.

\section{Convergence Analysis}\label{Methodology}

\begin{figure*}
\centering
\begin{tikzpicture}
\tiny
\node[draw]{
  \begin{tikzpicture}
    \draw[red, solid, thick] (0,0) -- (1,0) node[right, black]{FLamma};
    \draw[blue, dashed, thick] (3,0) -- (4,0) node[right, black]{FedAvg Baseline };
    \draw[green, densely dashed, thick] (6,0) -- (7,0) node[right, black]{FedProx Baseline};
    \draw[orange, dashed, thick] (9,0) -- (10,0) node[right, black]{ q-FFL Baseline };
    \draw[violet, dashed, thick] (12,0) -- (13,0) node[right, black]{Incentive Baseline};
  \end{tikzpicture}
};

\end{tikzpicture}

{\includegraphics[height=3.8cm, width=0.3\textwidth]{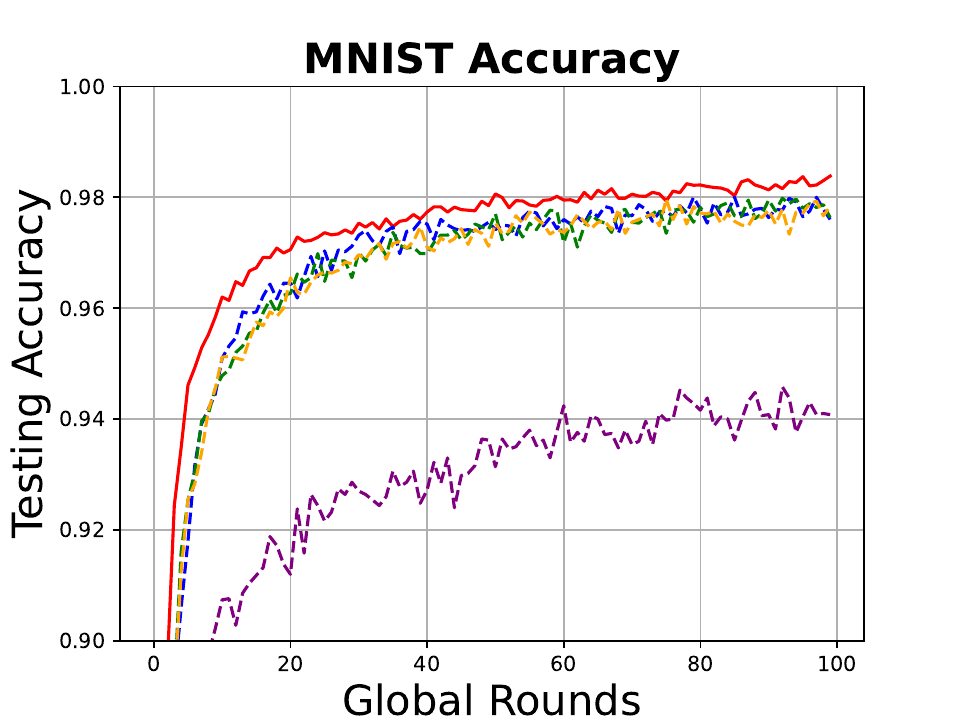}}
{\includegraphics[height=3.8cm, width=0.3\textwidth]{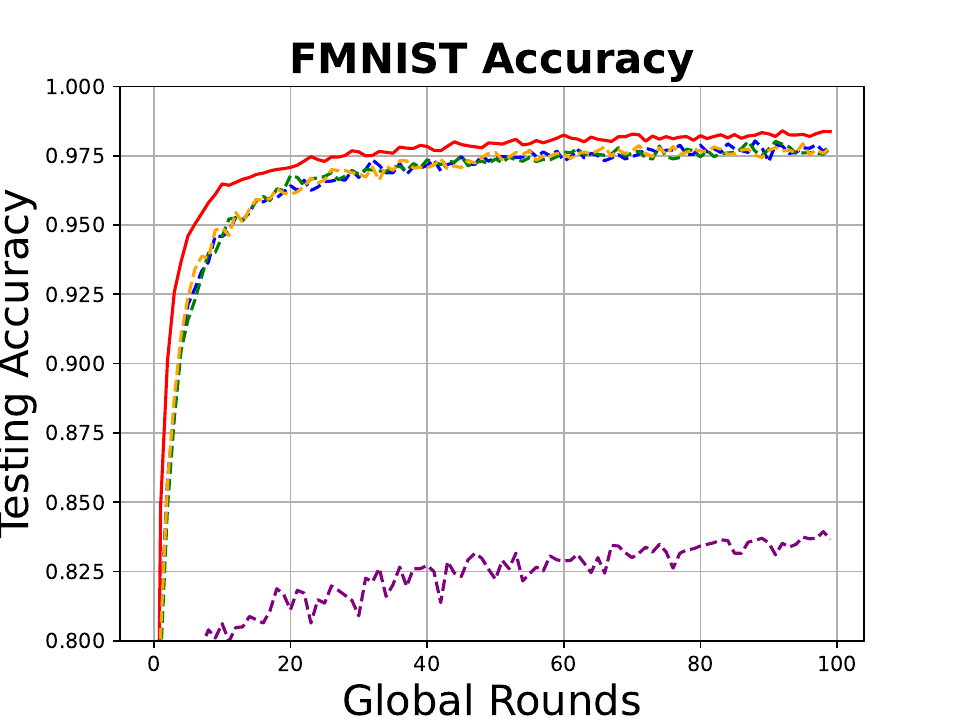}}
{\includegraphics[height=3.8cm, width=0.3\textwidth]{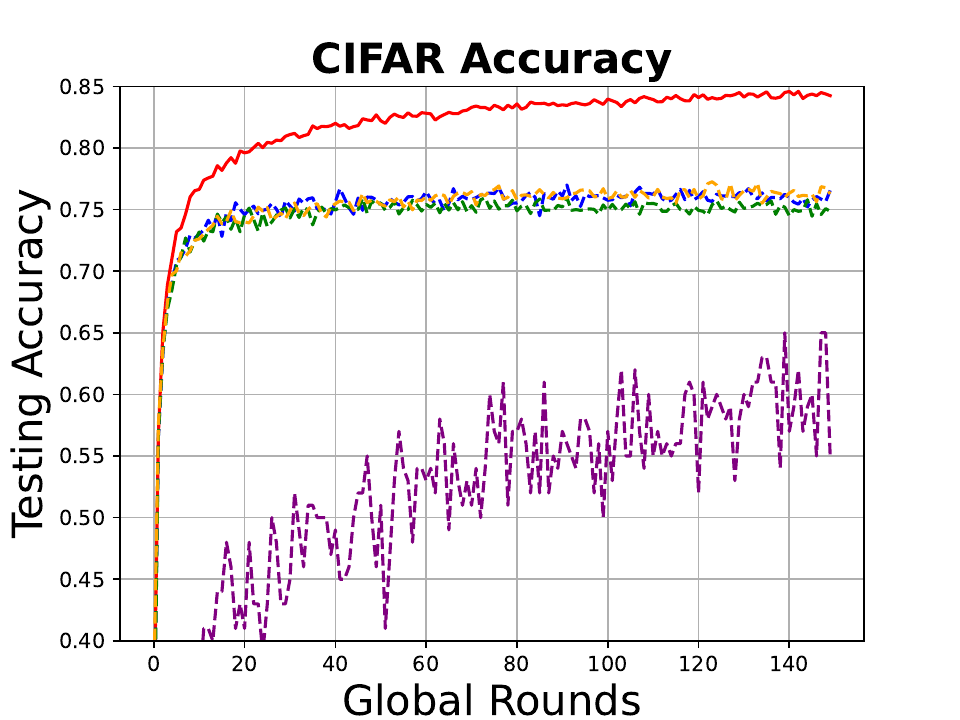}}
\\
\centering

{\includegraphics[height=3.8cm, width=0.3\textwidth]{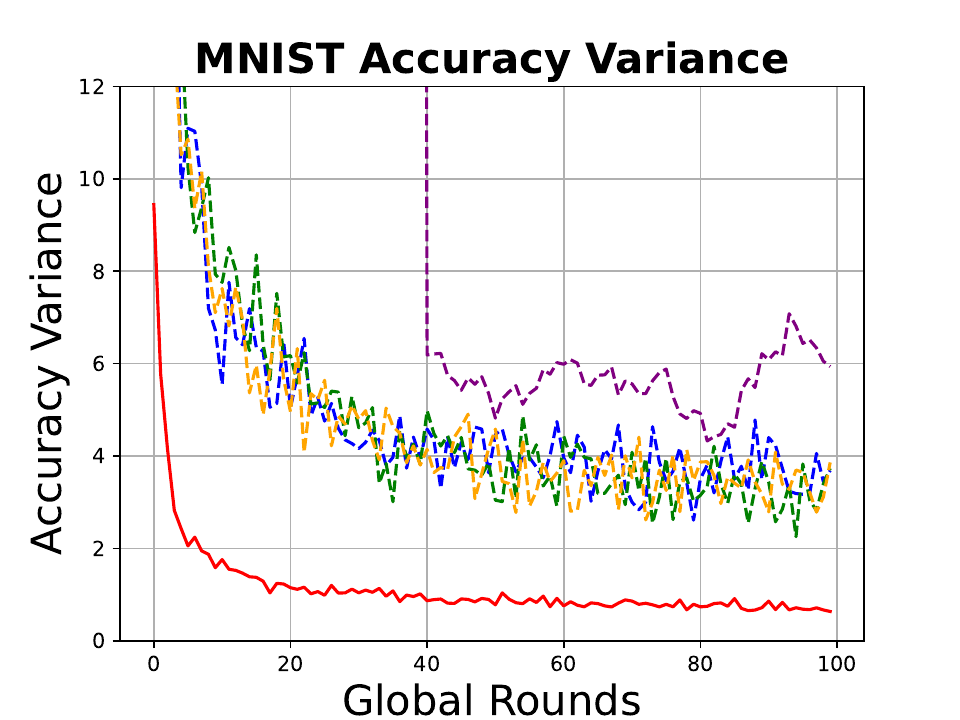}}
{\includegraphics[height=3.8cm, width=0.3\textwidth]{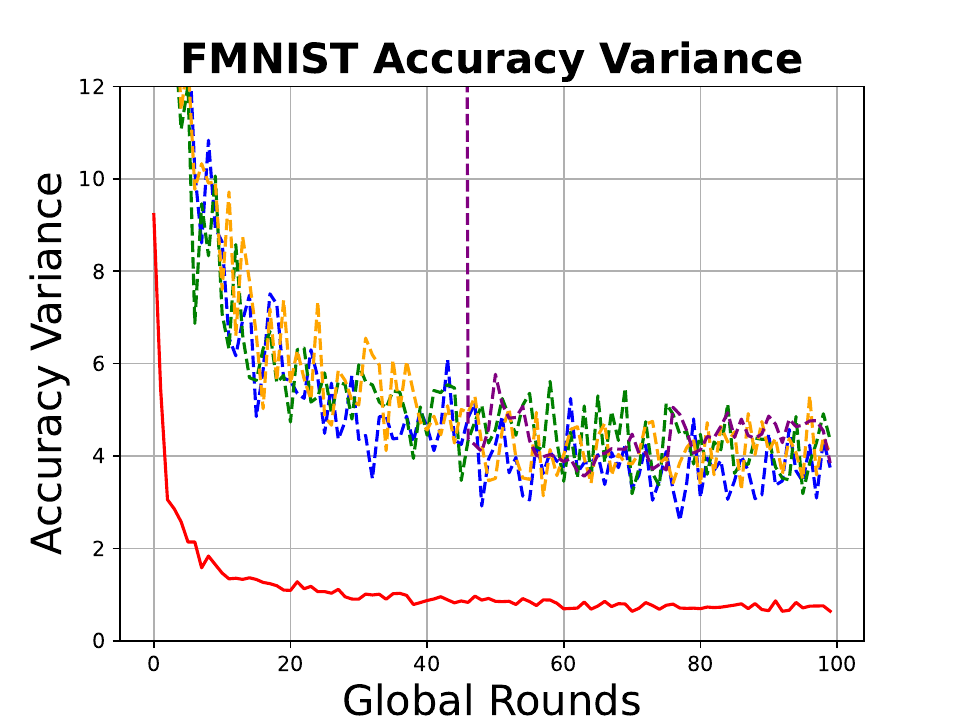}}
{\includegraphics[height=3.8cm, width=0.3\textwidth]{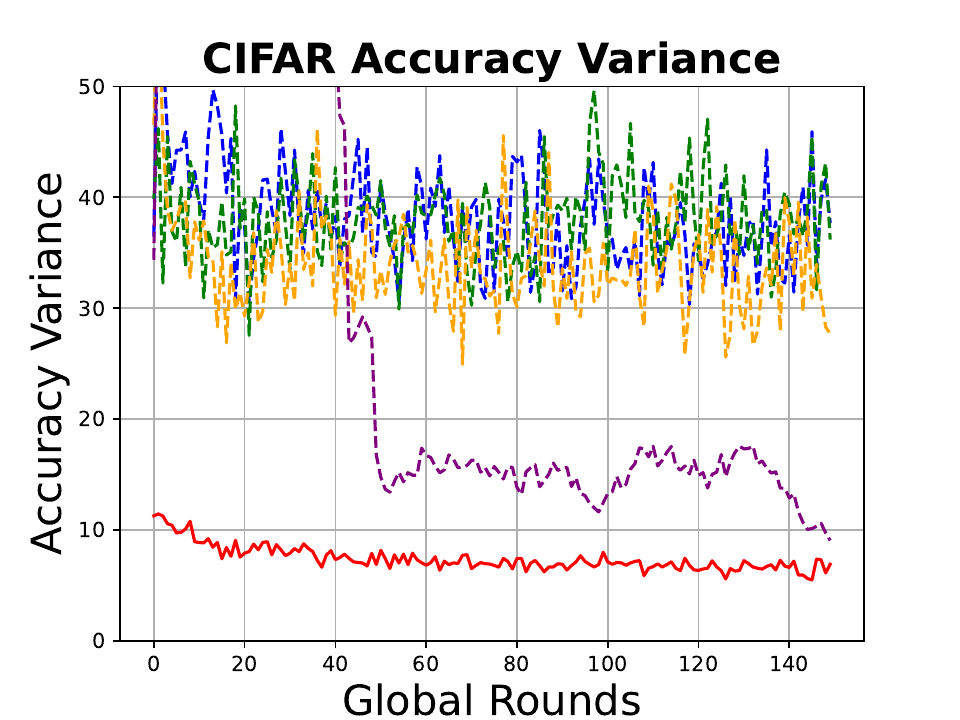}}
\\

\centering
\caption{Comparison of the FLamma with the baselines including FedAvg, FedProx, q-FFL, and Incentivization in terms of accuracy, and accuracy variance on IID dataset.}
\label{figure4}
\end{figure*}
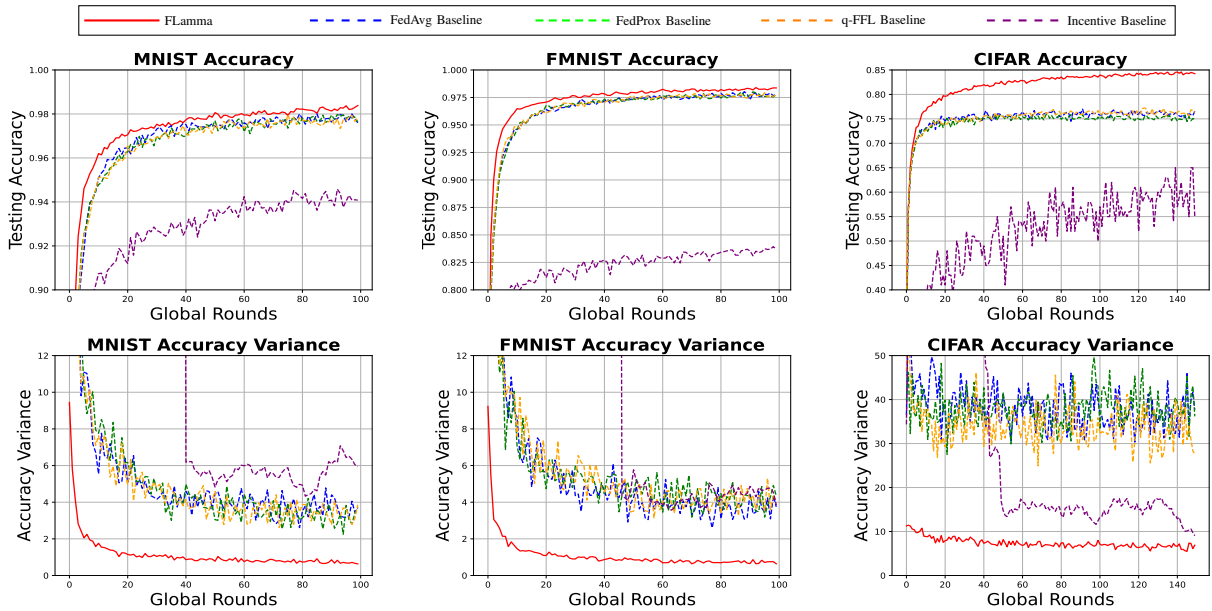
\begin{figure*}
\centering
{\includegraphics[height=3.8cm, width=0.3\textwidth]{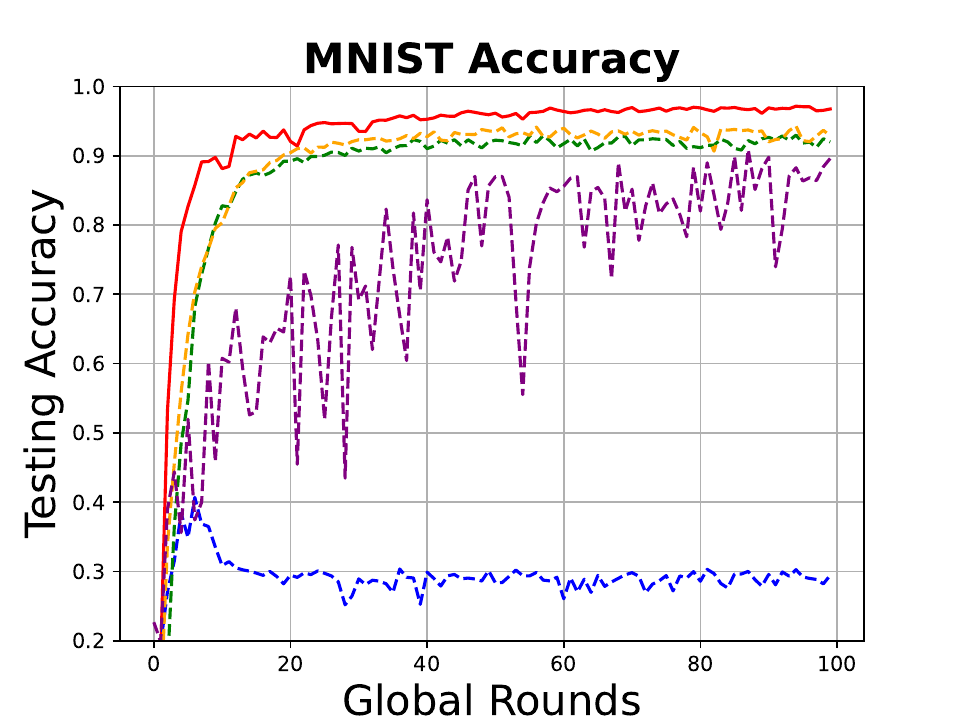}}
{\includegraphics[height=3.8cm, width=0.3\textwidth]{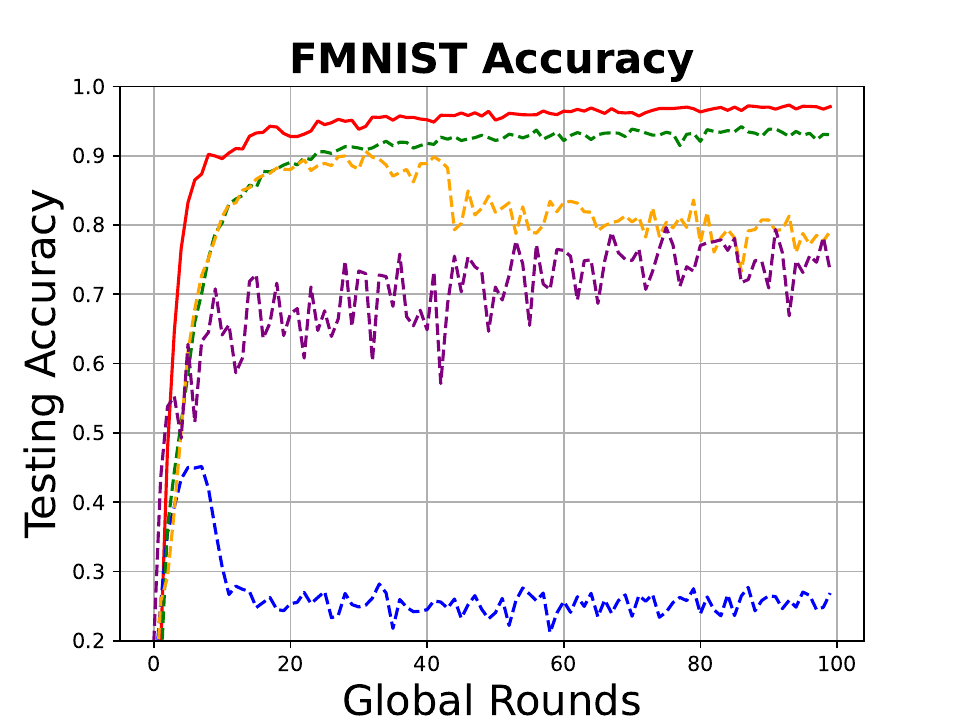}}
{\includegraphics[height=3.8cm, width=0.3\textwidth]{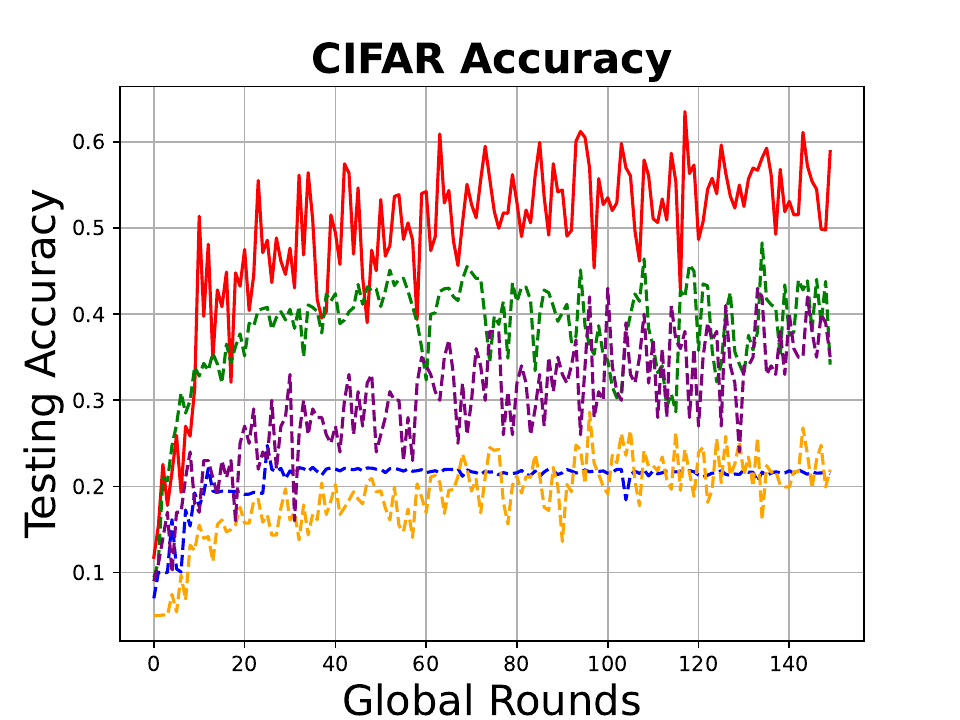}}
\\
\centering
{\includegraphics[height=3.8cm, width=0.3\textwidth]{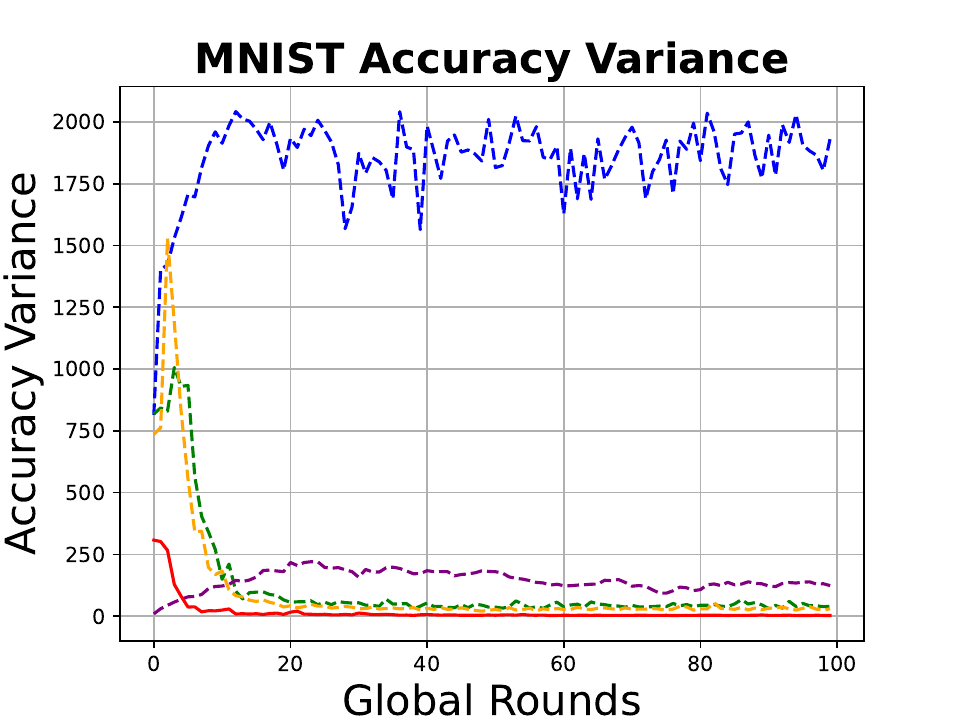}}
{\includegraphics[height=3.8cm, width=0.3\textwidth]{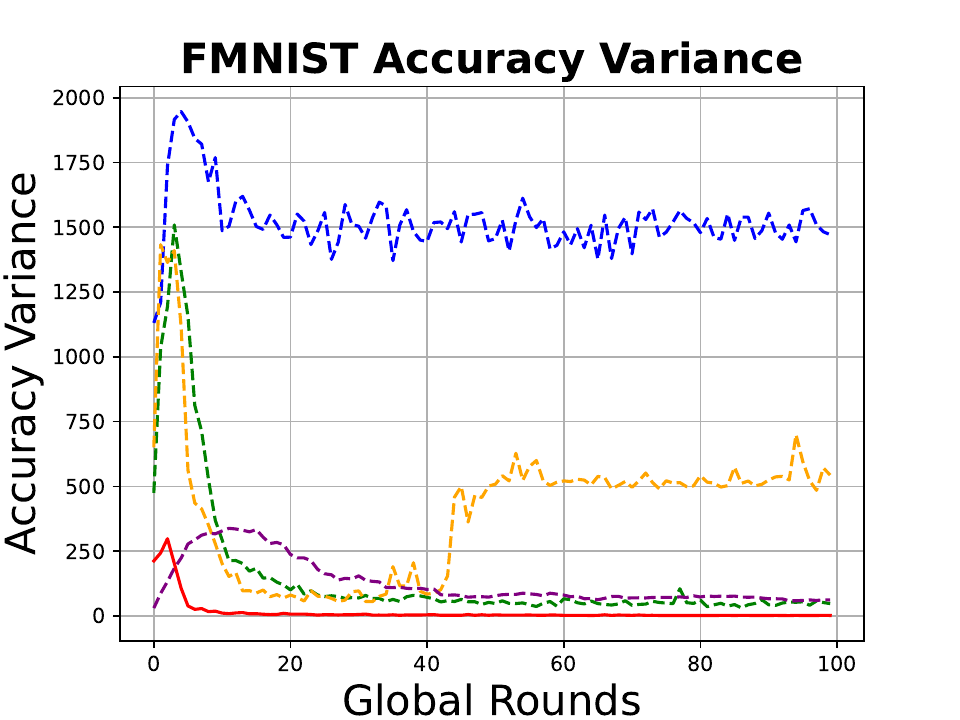}}
{\includegraphics[height=3.8cm, width=0.3\textwidth]{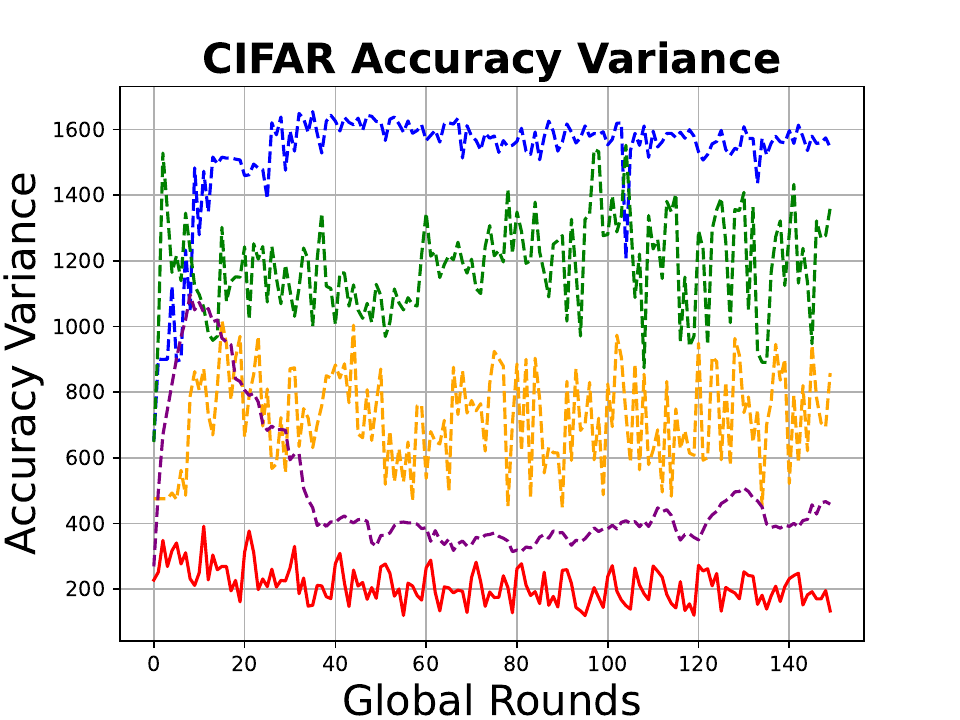}}
\caption{Comparison of the FLamma with the baselines including FedAvg, FedProx, q-FFL, and Incentivization in terms of accuracy, and accuracy variance on non-IID dataset.}
\label{figure5}
\end{figure*}

Using the FedAvg setting \cite{li2019convergence}, we have the following convergence analysis for our proposed setting with a decay factor.

\vspace{1em}

\noindent\textbf{\textit{Assumption 1.}}
We assume the following properties hold for all $i$:
% \begin{enumerate}

\noindent 1) Strong Convexity, i.e, $F_i(w)$ is \(\rho\)-strongly convex if it satisfies the following inequality: \[
    F_i(w') \geq F_i(w) + \langle \nabla F_i(w), w' - w \rangle + \frac{\rho}{2} \| w' - w \|^2,\; \forall \, w, w'.
    \]
2) $F_i(w)$ is $\beta$-smooth, i.e.,
\[
    \|\nabla F_i(w) - \nabla F_i(w')\| \leq \beta \|w - w'\|, \; \forall \; w, w'.
    \]
% \end{enumerate}

\noindent\textbf{\textit{Assumption 2.}} 
The variance of the stochastic gradients and the expected squared norm of the stochastic gradients in each device are bounded as follows:
\[
\mathbb{E} \left[ \|h_k(w_t^k) - \nabla F_k(w_t^k)\|^2 \right] \leq \sigma_k^2\; \text{and}\; \mathbb{E} \left[ \|h_k(w_t^k)\|^2 \right] \leq G^2,
\]
for all \(k = 1, \ldots, N\) and \(t = 1, \ldots, T-1\),
where $h_k(w_t^k)$ is the stochastic gradient of the local objective function on the $k$-th device.

% \noindent\textbf{\textit{Assumption 2.}} 
% The variance of stochastic gradients in each device is bounded:
% \[
% \mathbb{E} \left[ \|h_k(w_t^k) - \nabla F_k(w_t^k)\|^2 \right] \leq \sigma_k^2 \quad \text{for} \; k = 1, \ldots, N,
% \]
% where $h_k(w_t^k)$ is the stochastic gradient of the local objective function on the $k$-th device.

% % \vspace{1em}

% \noindent\textbf{\textit{Assumption 3.}}
% For all $k = 1, \ldots, N$ and $t = 1, \ldots, T-1$, the expected squared norm of stochastic gradients is uniformly bounded, i.e., $\mathbb{E} \left[ \|h_k(w_t^k)\|^2 \right] \leq G^2.$

% \vspace{1em}

\vspace{1em}
\noindent\textbf{\textit{Assumption 3.}} 
For $\eta > 0$, we have the following bound for the variance: $\mathbb{E} \left\| \mathbf{w}_t - \mathbf{w}_{t+1} \right\|^2 
\leq \frac{4}{K} \eta^2 \tau^2 G^2.$
\vspace{1em}

% \noindent\textbf{\textit{Assumption 4.}} Given Assumptions 1 and \(\eta \leq \frac{1}{4\beta}\), we have: $\mathbb{E} \|\mathbf{w}_{t+1} - w^*\|^2 \leq (1 - \eta \mu)\mathbb{E} \|\mathbf{w}_t - w^*\|^2 + \eta^2 \mathbb{E} \sum_{i=1}^N p_i \|\nabla F_i(w_t) - \nabla F(w_t)\|^2 + 6\beta\eta^2+ 2\eta \sum_{k=1}^N p_k \gamma \mathbb{E} \|\mathbf{w}_t - \mathbf{w}_k^t\|^2.$

\noindent\textbf{\textit {Assumption}}\textbf{\textit { 4.}} According to Assumption 2, the variance of the stochastic gradients is bounded as follows: $\mathbb{E} \sum_{i=1}^N p_i \gamma \|\nabla F_i(w_t) - \nabla F(w_t)\|^2 \leq \sum_{k=1}^N p_k^2 \gamma \sigma_k^2.$

\vspace{0.8em}
\noindent\textbf{\textit{Assumption 5.}} Given Assumption 2, the divergence between the local models and the global model is bounded: $\mathbb{E} \left[ \sum_{k=1}^N p_k \|\mathbf{w}_t - \mathbf{w}_k^t\|^2 \right] \leq 4\eta^2 (\tau_{max} - 1)^2 G^2.$
\vspace{1em}

\noindent\textbf{\textit{Theorem.}}
Given the above assumptions, $\tau_{\text{max}}$ the worst-case number of local epochs, $\kappa = \frac{\beta}{\rho},  \xi = \max\{8\kappa, \tau_{\text{max}}\}$, $\eta = \frac{2}{\rho \xi}$ and \(C = \frac{4}{K} \tau_{\text{max}}^2G^2\), the convergence bound is given by:
\begin{align}
 \mathbb{E}[F(\mathbf{w}_T)] - F^* &\leq \frac{\kappa}{T} \left( \frac{2(B+C)}{\rho} + \frac{\rho \xi \gamma_{\max} }{2} M \right), 
\label{eq2}
\end{align}
% + \frac{\kappa \rho \gamma_{max}}{\phi \beta},
% \; \phi = \xi + T - 1$
where $M =\mathbb{E}\|\mathbf{w}_1 - \mathbf{w}^*\|^2$,  and $B =\frac{1}{\rho}\sum_{k=1}^N p_k^2 \gamma_{\max} \sigma_k^2 + 6\beta\eta^2+ 8(\tau_{\text{max}} - 1)^2 G^2.$ 

% 8(\tau_{\text{max}} - 1)^2 G^2.$

% \noindent\textit{Proof.} 
% From Assumption 5, the recurrence relation for one-step updates is given by: $\mathbb{E} \|\mathbf{w}_{t+1} - w^*\|^2 \leq (1 - \eta \mu)\mathbb{E} \|\mathbf{w}_t - w^*\|^2 
% + \eta^2 \sum_{i=1}^N p_i \|\nabla F_i(w_t) - \nabla F(w_t)\|^2 + 6L \eta^2 + 2\eta \sum_{k=1}^N p_k \gamma \mathbb{E} \|\mathbf{w}_t - \mathbf{w}_k^t\|^2.$ Using Assumption 6, the variance term can be bounded as: $\mathbb{E} \sum_{i=1}^N p_i \gamma \|\nabla F_i(w_t) - \nabla F(w_t)\|^2 \leq \sum_{k=1}^N p_k^2 \gamma \sigma_k^2.$ Substituting this into the recurrence relation gives: $\mathbb{E} \|\mathbf{w}_{t+1} - w^*\|^2 \leq (1 - \eta \mu)\mathbb{E} \|\mathbf{w}_t - w^*\|^2 + \eta^2 \sum_{k=1}^N p_k^2 \gamma \sigma_k^2 + 6L \eta^2 + 2\eta \sum_{k=1}^N p_k \gamma \mathbb{E} \|\mathbf{w}_t - \mathbf{w}_k^t\|^2.$

\noindent\textit{Proof.} Given Assumptions 1 and 5, we start by using the recurrence relation for one-step updates:
\[\mathbb{E} \|\mathbf{w}_{t+1} - w^*\|^2 \leq (1 - \eta \rho)\mathbb{E} \|\mathbf{w}_t - w^*\|^2 +\]\vspace{-1em}\[
\eta^2 \mathbb{E} \sum_{i=1}^N p_i \|\nabla F_i(w_t)-\nabla F(w_t)\|^2 +  6\beta\eta^2+\]\vspace{-1em}\[2\eta \sum_{k=1}^N p_k \gamma \mathbb{E} \|\mathbf{w}_t - \mathbf{w}_k^t\|^2.
 \]
 \noindent Using Assumption 5 and substituting it into the previous equation gives: $\mathbb{E} \|\mathbf{w}_{t+1} - w^*\|^2 \leq (1 - \eta \rho)\mathbb{E} \|\mathbf{w}_t - w^*\|^2 + $$\eta^2 \sum_{k=1}^N p_k^2 \gamma \sigma_k^2 + 6\beta\eta^2 + 8\eta^3 (\tau_{\text{max}} - 1)^2 G^2.$ Next, using the smoothness property of \(F(w)\), we can bound the difference between the objective function at the \(T\)-th iteration and the optimal value: $F(w_T) - F^* \leq \frac{\beta}{2} \mathbb{E} \|\mathbf{w}_T - w^*\|^2.$ Summing this recurrence relation over \(t = 1\) to \(T\), and applying the resulting bounds, we arrive at:
$F(w_T) - F^* \leq \frac{\beta}{2 \rho T} \Big( \mathbb{E} \|\mathbf{w}_1 - w^*\|^2 + T \eta^2 \sum_{k=1}^N p_k^2 \gamma \sigma_k^2$  $+ 6\beta T \eta^2 + 8 \eta^3 T (\tau_{\text{max}} - 1)^2 G^2 \Big).$

\noindent Taking the expectation of both sides, then incorporating these derived components and applying the assumptions, we rearrange the terms to yield the final result, as expressed in Eq.~(\ref{eq2}). 

In our method, we incorporate both a decay factor \(\gamma\) and the number of local epochs \(\tau_i\) for each client. In the \textit{worst-case scenario} where \(\gamma_{\max} = 1\), client contributions are unscaled by \(\gamma\), particularly in early rounds or when local updates are small. As training progresses, either with increasing rounds \(t\) or larger \(\tau_i\), \(\gamma\) decays, reducing the impact of clients with many local updates. This mechanism tightens the convergence bound by limiting the influence of clients with high \(\tau_i\), ensuring stability as \(\gamma\) approaches zero. Clients can select \(\tau_i\) within a bounded range \([ \tau_{\text{min}}, \tau_{\text{max}} ]\), which prevents excessive divergence between local and global models. Even in the worst case where \(\tau_{\text{max}}\) is large, the decay factor \(\gamma_{\max}\) compensates by shrinking, limiting the contribution of clients with high local updates. As a result, \(\gamma\) ensures the bound remains tight throughout training, behaving similarly to FedAvg, with only a marginal increase due to the dynamic scaling of client contributions.

\begin{table*}[!htbp]
% \LARGE
\caption{ Accuracy and Accuracy Variance for IID and non-IID Datasets across CIFAR10, and FMNIST, and MNIST}
\centering

\renewcommand{\arraystretch}{0.75}
\tiny

% \resizebox{\textwidth}{!}{%
\scalebox{1.8}{
\begin{tabular}{|l|c|c|c|c|}
   \multicolumn{5}{c}{\textbf{CIFAR10 Dataset}} \\\hline
   \textbf{Algorithm} & \textbf{Accuracy (IID)} & \textbf{Variance (IID)} & \textbf{Accuracy (Non-IID)} & \textbf{Variance (Non-IID)} \\\hline
   FedAvg   & 75.30 $\pm$ 0.25 & 43.22 $\pm$ 1.30 & 22.48 $\pm$ 0.87 & 1615.00 $\pm$ 3.24 \\
   FedProx  & 75.11 $\pm$ 0.18 & 36.18 $\pm$ 1.15 & 33.03 $\pm$ 0.75 & 1184.55 $\pm$ 4.67 \\
   q-FFL    & 75.60 $\pm$ 0.21 & 27.12 $\pm$ 1.23 & 25.56 $\pm$ 0.88 & 720.87 $\pm$ 2.18 \\
   Incentivization    & 57.05 $\pm$ 0.98 & 9.04 $\pm$ 0.89 & 40.44 $\pm$ 1.03 & 458.44 $\pm$ 1.95 \\
   FLamma   & \textbf{84.36} $\pm$ 0.32 & \textbf{7.52} $\pm$ 0.83 & \textbf{56.47} $\pm$ 0.97 & \textbf{241.79} $\pm$ 1.12 \\\hline
\end{tabular}

}

\vspace{1em}

% \resizebox{\textwidth}{!}{%
\scalebox{1.8}{
\begin{tabular}{|l|c|c|c|c|}
   \multicolumn{5}{c}{\textbf{FMNIST Dataset}} \\\hline
   \textbf{Algorithm} & \textbf{Accuracy (IID)} & \textbf{Variance (IID)} & \textbf{Accuracy (Non-IID)} & \textbf{Variance (Non-IID)} \\\hline
   FedAvg   & 98.19 $\pm$ 0.15 & 5.64 $\pm$ 0.72 & 26.51 $\pm$ 0.75 & 1521.64 $\pm$ 2.54 \\
   FedProx  & 97.95 $\pm$ 0.18 & 5.82 $\pm$ 0.57 & 94.11 $\pm$ 0.95 & 155.82 $\pm$ 2.28 \\
   q-FFL    & 98.14 $\pm$ 0.22 & 6.07 $\pm$ 0.61 & 77.96 $\pm$ 0.93 & 406.46 $\pm$ 2.03 \\
   Incentivization    & 84.36 $\pm$ 0.54 & 3.84 $\pm$ 0.65 & 74.82 $\pm$ 0.81 & 62.10 $\pm$ 1.47 \\
   FLamma   & \textbf{98.84} $\pm$ 0.19 & \textbf{1.06} $\pm$ 0.43 & \textbf{97.29} $\pm$ 0.98 & \textbf{13.88} $\pm$ 0.65 \\\hline
\end{tabular} }
\vspace{1em}

% \resizebox{\textwidth}{!}{%
\scalebox{1.8}{
\begin{tabular}{|l|c|c|c|c|}
   \multicolumn{5}{c}{\textbf{MNIST Dataset}} \\\hline
   \textbf{Algorithm} & \textbf{Accuracy (IID)} & \textbf{Variance (IID)} & \textbf{Accuracy (Non-IID)} & \textbf{Variance (Non-IID)} \\\hline
   FedAvg   & 98.15 $\pm$ 0.24 & 5.44 $\pm$ 0.95 & 34.98 $\pm$ 0.92 & 1829.76 $\pm$ 3.82 \\
   FedProx  & 98.25 $\pm$ 0.36 & 5.54 $\pm$ 0.91 & 93.34 $\pm$ 0.76 & 154.58 $\pm$ 1.43 \\
   q-FFL    & 98.33 $\pm$ 0.27 & 5.21 $\pm$ 0.78 & 93.24 $\pm$ 0.83 & 148.23 $\pm$ 1.84 \\
    Incentivization    & 94.2 $\pm$ 0.64 &  5.93 $\pm$ 0.88 & 88.28 $\pm$ 0.78 & 123.47 $\pm$ 1.25 \\
   FLamma   & \textbf{98.74} $\pm$ 0.16 & \textbf{1.12} $\pm$ 0.67 & \textbf{97.62} $\pm$ 0.48 & \textbf{48.51} $\pm$ 0.59 \\\hline
\end{tabular}
}
\label{table}
\end{table*}

\section{Experiments and Results}\label{experimental}

This section provides the specifics of the evaluation and performance results and a discussion of the proposed method. The evaluation specifics include the dataset details, FL models and strategies, and experiment setup details. Performance results are depicted for (1) accuracy under three baseline FL strategies and the proposed FLamma strategy and (2) accuracy variance under the same three baseline FL strategies and the proposed strategy for both IID and non-IID dataset splits.

\subsection{Setup}\label{evaluation}

% \noindent\textbf{\textit{Dataset Details.}} 
% % We evaluate the performance of FLamma, using CNN models \cite{fukushima1980neocognitron} including ResNet-18 \cite{he2016deep} and LeNet-5 \cite{lecun1989backpropagation} on three different popular benchmark datasets including MNIST \cite{deng2012mnist}, FashionMNIST \cite{xiao2017fashion}, and CIFAR10 \cite{krizhevsky2009learning} on IID datasets.

% this is new

\noindent\textbf{\textit{Datasets.}} We evaluate the performance of FLamma using convolutional neural network (CNN) models, specifically ResNet-18 \cite{he2016deep} and LeNet-5 \cite{lecun1989backpropagation}, across three widely used benchmark datasets: MNIST \cite{deng2012mnist}, FashionMNIST \cite{xiao2017fashion}, and CIFAR10 \cite{krizhevsky2009learning}, all under independent and identically distributed (IID) and non-IID conditions.

To simulate non-IID data in our FL setup, we use a shard-based method. The dataset is partitioned into smaller shards, which are distributed unevenly among clients, ensuring each receives data that does not fully represent the overall distribution, mimicking real-world scenarios where client data varies significantly.

We first divide the dataset into more shards than there are clients to maintain diversity. The data is sorted by labels so that each shard contains primarily samples from a single class. Shards are then randomly assigned to clients, with each receiving a few, creating a biased distribution. This method results in clients having imbalanced and class-specific data, reflecting the variability typically observed in federated learning environments.

% \vspace{1em}
\vspace{0.6em}
\noindent\textbf{\textit{FL Baselines.}} We compared the performance of FLamma against baselines FedAvg \cite{mcmahan2017communication}, FedProx \cite{li2020federated}, q-FFL \cite{li2019fair} and Incentivization \cite{wu2023incentive} methods to validate its effectiveness. These baselines are chosen for their relevance to key federated learning challenges. 

\begin{itemize}
% \cite{xu2021gradient}

\item \textbf{FedAvg:} This baseline serves as the simplest baseline with random client selection without any specialized mechanism to address client heterogeneity.

\item \textbf{FedProx:} FedProx enhances the FedAvg algorithm by incorporating a proximal term into the objective function. This adjustment helps address the challenges posed by non-IID data distributions, commonly encountered in federated learning.

\item \textbf{q-FFL}: This method aims to enhance fairness and reduce accuracy variance by assigning different weights to clients based on their empirical losses. By focusing on a more balanced distribution of accuracy among heterogeneous model owners, it addresses disparities and promotes fairness within federated learning environments.

Different from these baselines, our method introduces a dynamic, game-theoretic approach where the server adjusts a decay factor over time, balancing client contributions. Unlike FedAvg, which lacks mechanisms for fairness or client heterogeneity, and FedProx, which only addresses heterogeneity without fairness considerations, our approach balances both fairness and heterogeneity. While q-FFL directly addresses fairness by weighting clients with higher losses, our method promotes long-term fairness by dynamically adjusting client influence over time, ensuring both fairness and improved system efficiency through adaptive contribution scaling.

\item \textbf{Incentive Baseline:} This approach evaluates each client’s marginal contribution to the global model, prioritizing clients whose updates improve accuracy. In contrast, our approach incorporates a dynamic decay factor within a Stackelberg game framework, balancing both fairness and contributions over time to promote sustained client participation.

\end{itemize}
% \vspace{1em}

\subsection{Evaluation Metrics}
The evaluation focuses on two key performance metrics: Testing accuracy and accuracy variance, which are crucial for assessing both the effectiveness and fairness of the FL system.

\vspace{1em}
\noindent\textbf{\textit{Testing Accuracy.}} Testing accuracy measures the overall performance of the global model after aggregating local models, expressed as the percentage of correct predictions. In FL, the goal is to achieve high global accuracy while maintaining fairness and consistency across all clients.
\vspace{1em}

\noindent\textbf{\textit{Accuracy Variance.}}
Accuracy variance measures the inconsistency in performance across different clients. A lower accuracy variance indicates that all clients achieve similar accuracy levels, reflecting fairness in the learning process. In FL, high accuracy variance is undesirable, as it suggests that some clients contribute disproportionately or experience significantly lower accuracy than others, which can result in unfairness. Our aim with FLamma is to minimize this variance, especially in non-IID settings where client data is highly heterogeneous.

% We evaluate our experimental results against several federated learning baselines, including FedAvg \cite{mcmahan2017communication}, FedProx \cite{li2020federated}, which addresses heterogeneity by adding a proximal term to local updates, q-FFL \cite{li2019fair}, which promotes fairness by weighting clients' losses to reduce accuracy variance.

% CFFL \cite{lyu2020collaborative}, which enforces fairness through reputation-based convergence to different models, and CGSV \cite{xu2021gradient}, which assesses the marginal contribution of each client's update.
\vspace{0.6em}
\noindent\textbf{\textit{Details of Experimental Setup}}.
We conducted our experiments in comparison with the aforementioned FL methods: FedAvg, FedProx, q-FFL and Incentivization. Each experiment was run 3 times to ensure correctness and accuracy. The experiments were executed on a server equipped with an NVIDIA GeForce RTX 3080Ti GPU, Intel(R) Core(TM) i9-109000X CPU, and 64G of RAM. We run both the server and clients on the same machine in a simulated environment where the clients train their local model for a number of local iterations on their own data. Once they complete their local iterations, they communicate their updated weights to the server along with other important information such as loss and accuracies. This configuration is supported by the fact that the performance metrics we evaluate are independent of the physical separation between the server and clients. 

In our proposed setting, the clients train for ten local iterations. The server then assigns contributions to each of the clients based on their performance in the ten local iterations they were assigned. The server aggregates the weights and updates its global model. The server then assigns contributions (a value between 0 and 1) to the clients based on their weights' Euclidian distance from the global model weights. Those clients with weights that are closer to the global model's weights will be assigned a higher contribution value while clients with weights which are farther from the global model's weights will be assigned a lower contribution. The contribution thereafter is updated every 10 global rounds. This updating process dictates the selection of clients throughout the global training process. As clients get selected and contribute to the training, the decay factor gets applied, reducing the contribution of the clients over time between the tenth round updates. LeNet-5 was trained on MNIST and FMNIST datasets, while ResNet-18 was trained on CIFAR10 dataset.

%  Our experiments were conducted using the FedAvg. We ran each experiment three times. We executed all experiments on a server, which is equipped with NVIDIA GeForce RTX 3080Ti GPU, Intel(R) Core(TM) i9-10900X CPU, and 64G RAM. We run both the server and clients on the same machine, a configuration supported by the fact that the performance metrics we evaluate are independent of the physical separation between the server and clients.

% In the initial round, clients were trained for one epoch. Clients were then selected based on their performance in this first training round. Following this, the contribution of each client was updated every ten rounds. This updating process informed the selection of clients for the subsequent set of rounds. LeNet-5 was trained on FMNIST and MNIST datasets, while ResNet-18 was trained on CIFAR10 dataset.   

\subsection{Performance Results and Discussion}

In this section, we present the evaluation results of our proposed method, FLamma, compared to several baseline methods, including FedAvg, FedProx, q-FFL and Incentivization. The performance of each method is evaluated across different datasets, including CIFAR10, MNIST, and FMNIST, under both IID (Independent and Identically Distributed) and non-IID conditions.

Figs. \ref{figure4} (IID) and \ref{figure5} (non-IID), demonstrates the results of the method. The top row of Figs. \ref{figure4} and \ref{figure5} present the global testing accuracy over the course of the global training rounds for each method across the datasets. The Y-axis represents the testing accuracy, ranging from 0.0 to 1.0, while the X-axis shows the number of global training rounds, with a maximum of 100 rounds for MNIST and FMNIST, and 150 rounds for CIFAR10. 

The bottom row of Figs. \ref{figure4} and \ref{figure5} illustrate the accuracy variance across clients during global training. The Y-axis measures the variance in accuracy (with lower values indicating more uniform performance across clients), and the X-axis corresponds to the number of global training rounds. Across all datasets, FLamma achieves substantially lower accuracy variance compared to the baseline methods, as indicated by the consistently lower red curve. This reduction in variance is notable in all datasets, where FLamma rapidly reduces the variance in early rounds and maintains it at a low level throughout training.
The lower accuracy variance achieved by FLamma reflects its ability to ensure fairness across clients, as the model performs more uniformly regardless of the heterogeneity in clients' data distributions. In contrast, the baselines exhibit higher variance, indicating less consistency in client performance.

% \vspace{1em}

\vspace{0.6em}
\noindent\textbf{\textit{Performance under IID Data Split.}} In the IID setting (Fig. \ref{figure4}), FLamma demonstrates substantial improvements in reducing accuracy variance compared to baseline methods, reflecting more consistent performance across clients, which is essential for ensuring fairness. In particular, as shown in Table \ref{table}, in the CIFAR10 dataset, FLamma shows a clear advantage, improving accuracy  by up to 11.59\% compared to q-FFL (75.60\% to 84.36\%) and reduces variance by 82.60\% compared to FedAvg (43.22 to 7.52). In both MNIST and FMNIST, FLamma converges to similarly high accuracy levels while significantly reducing accuracy variance, ensuring fairer performance across all clients. This demonstrates that FLamma is not only effective at maintaining competitive accuracy but also excelling at promoting fairness by minimizing performance disparities among clients, even when data is uniformly distributed.

% For instance, in the CIFAR10 dataset, FLamma achieves an accuracy of 84.36\% with an accuracy variance of 6.52, outperforming FedAvg (75.30\% accuracy, 38.22 variance). Across all datasets, FLamma (represented by the red curve) consistently achieves either higher or comparable global accuracy relative to FedAvg, FedProx, and q-FFL. In particular, in the CIFAR-10 dataset, FLamma shows a clear advantage, improving accuracy as training progresses. In both MNIST and FMNIST, FLamma converges to similarly high accuracy levels while significantly reducing accuracy variance, ensuring fairer performance across all clients. This demonstrates that FLamma is not only effective at maintaining competitive accuracy but also excels at promoting fairness by minimizing performance disparities among clients, even when data is uniformly distributed.
\vspace{1em}

\noindent\textbf{\textit{{Performance under Non-IID Data Split.}}} In the more challenging non-IID scenario (Fig. \ref{figure5}), FLamma demonstrates even greater advantages over baseline methods. Traditional approaches like FedAvg and FedProx often struggle with high accuracy variance due to the heterogeneity of client data. However, FLamma effectively addresses this issue, significantly reducing accuracy variance while maintaining high global accuracy. As mentioned in Table \ref{table}, in CIFAR10 dataset, FLamma achieves an impressive 120.93\% improvement in accuracy over q-FFL (25.56\% to 56.47\%) and reduces variance by 85.03\% compared to FedAvg (1615.00 to 241.79). Similar trends are observed in the MNIST and FMNIST datasets. In FMNIST dataset, FLamma achieves an impressive 24.79\% improvement in accuracy over q-FFL (77.96\% to 97.29\%) and reduces variance by 99.09\% compared to FedAvg (1521.64 to 13.88). Similarly in MNIST, FLamma consistently delivers superior results in terms of both accuracy and fairness, maintaining a balance between client performance despite data heterogeneity. These results highlight FLamma's robustness in handling non-IID data, where it consistently improves fairness by reducing performance disparities across clients without sacrificing global accuracy.
\vspace{1em}

As shown in our experiments, FLamma has significant improvements in reducing accuracy variance compared to baseline methods. This reduction in variance ensures that individual clients have more consistent performance, which is critical for ensuring fairness across the federated system. Importantly, FLamma achieves these fairness improvements without compromising global accuracy. In fact, FLamma consistently achieves competitive or superior global accuracy compared to FedAvg, FedProx, q-FFL and Incentivization across various datasets.
% For example, in the CIFAR10 dataset under non-IID conditions, FLamma achieves an accuracy of 56.47\%, significantly outperforming FedProx (33.03\%) and q-FFL (25.56\%), with a much lower accuracy variance of 241.79. 
% Similar trends are observed in the MNIST and FMNIST datasets, where FLamma consistently delivers superior results in terms of both accuracy and fairness, maintaining a balance between client performance despite data heterogeneity. These results highlight FLamma's robustness in handling non-IID data, where it consistently improves fairness by reducing performance disparities across clients without sacrificing global accuracy.

\vspace{1em}

\section{Limitation and Future Work}

Despite promoting fairness and efficiency in our work, several limitations need to be addressed in future research. One limitation is the reliance on precise measurements of client contributions, which can be challenging in real-world scenarios where data quality and client capabilities vary widely. Additionally, while the incorporation of a decay factor into the assignment strategy promotes fairness and convergence over time, the optimal tuning of this decay factor requires non-negligible experimentation and may not be straightforward in diverse environments. Furthermore, there is no one decay factor that works for all learning tasks, implying the decay factor should be fine-tuned to each federated learning task on different datasets. Future research could also explore other techniques, such as reinforcement learning, to dynamically optimize hyperparameters and improve the approach's adaptability. 

\section{Conclusion}
In this paper, we proposed an FL framework inspired by Stackelberg game theory, which dynamically assigns local training epochs and regulates client contributions through a decay factor. Our approach, FLamma, leverages the hierarchical structure of the Stackelberg game, where the server acts as the leader, and clients respond as followers, optimizing their participation based on the server's decay factor. By incorporating the decay factor into the assignment strategy, our method not only adapts to evolving client contributions over time but also reduces the influence of clients' contributions, preventing them from dominating the learning process. This ensures that underperforming clients are not disproportionately underrepresented while maintaining overall system fairness. The key innovation of our method is the dynamic allocation of both local epochs and the decay factor to incentivize clients to maximize their contributions without allowing any client to dominate the updates. This prevents overrepresentation by some clients and ensures a more balanced participation across all clients, promoting fairness without undermining the overall learning process. Our experimental results demonstrate that FLamma significantly reduces the variance in accuracy among clients compared to baseline methods.% such as FedAvg, FedProx, q-FFL, and other incentivization methods.

\balance
% \bibliographystyle{IEEEtran}
% \bibliography{Bib} 

\end{document}